\documentclass[letterpaper, 10 pt, conference]{ieeeconf}
\usepackage{cite}
\usepackage{commath}
\usepackage{amsmath}
\usepackage{amssymb}
\usepackage{graphicx}
\usepackage[caption=false, font=footnotesize]{subfig}
\usepackage[misc]{ifsym}
\usepackage{epstopdf}
\usepackage{algorithm}
\usepackage{algorithmic}
\usepackage{siunitx}
\usepackage{dblfloatfix}

\usepackage{mathtools}
\usepackage{gensymb}
\usepackage{nicefrac}
\usepackage{xspace}
\usepackage{array}
\usepackage{makecell}

\usepackage{titlesec}

\IEEEoverridecommandlockouts
\title{A Biologically Inspired Global Localization System for Mobile Robots\\ Using LiDAR Sensor}

\author{
	\normalsize{
	Genghang Zhuang$^1$, 
	Carlo Cagnetta$^1$, 
	Zhenshan Bing$^1$, 
	Kai Huang$^2$, and 
	Alois Knoll$^1$ 
	}
	\thanks{\footnotesize{$^1$Department of Informatics, Technical University of Munich.}}
	\thanks{\footnotesize{$^2$School of Computer Science and Engineering, Sun Yat-sen University.}}
}

\begin{document}

\def\LiDAR{Li\textsc{dar}\xspace}
\def\LiDARs{Li\textsc{dar}s\xspace}

\def\figurename{Fig.}
\def\tablename{TABLE}

\DeclarePairedDelimiter\ceil{\lceil}{\rceil}
\DeclarePairedDelimiter\floor{\lfloor}{\rfloor}

\maketitle

\thispagestyle{empty}
\pagestyle{empty}

\begin{abstract}

Localization in the environment is an essential navigational capability for animals and mobile robots.
In the indoor environment, the global localization problem remains challenging to be perfectly solved with probabilistic methods.
However, animals are able to instinctively localize themselves with much less effort.
Therefore, it is intriguing and promising to seek biological inspiration from animals.
In this paper, we present a biologically-inspired global localization system using a \LiDAR sensor that utilizes a hippocampal model and a landmark-based re-localization approach.
The experiment results show that the proposed method is competitive to Monte Carlo Localization, and the results demonstrate the high accuracy, applicability, and reliability of the proposed biologically-inspired localization system in different localization scenarios.

\end{abstract}

\IEEEpeerreviewmaketitle


\section{Introduction}

As essential prerequisites for most autonomous driving tasks, robot localization and position tracking based on sensory data have been extensively investigated~\cite{fox1998position,thrun2002probabilistic}.
Global Navigation Satellite System (GNNS) is generally effective for the outdoor environment due to the reliable external reference sources for global localization.
However, in the indoor environment, GNNS-based localization is much less robust given the weaker accessibility of the satellites.
Existing research on global localization for the indoor environment extensively relies on probabilistic models to cope with the uncertainty of the environment and sensors.
Probabilistic methods for localization become dominant in indoor applications by offering a relatively robust and generic solution for the localization problem based on robot perception~\cite{thrun2002probabilistic}.
Characterized and distinguished by the problem being solved, probabilistic models such as Extended Kalman Filter (EKF) are utilized to tackle position tracking~\cite{thrun2002probabilistic,kalman1960new}, and Monte Carlo Localization (MCL)~\cite{dellaert1999monte} focuses on the global localization problem.
However, probabilistic methods can entail the robot system a high computational overhead, and the runtime performance can be subsequently subject to the computational resources on a power-constrained robot.

On the other hand, without the need for high computational resources, many animals are naturally born with powerful and efficient navigation skills, such as the echolocation by bats, magnetic-based location by migratory birds, and also the most common vision-based navigation.
Although animals only reply on a collection of imprecise and imperfect perception, their various navigation skills yet still outperform state-of-the-art artificial technologies.
Studies from neuroscience have revealed that the hippocampus in the brain plays an important role in spatial navigation by coordinating several types of neurons with different functionalities, such as head direction cells, grid cells, and place cells~\cite{moser2008place}.
Therefore, it is intriguing and can be promising to seek inspiration from the navigation system of animals.

Inspired by the above neuroscience findings, a number of biologically inspired methods have been proposed to solve spatial mapping tasks by building not strictly Cartesian but topologically consistent representations for the environments
~\cite{milford2004ratslam,milford2008mapping,steckel2013batslam}.
However, the global localization problem is rarely investigated based on biologically inspired models.
In addition, most of the existing biologically inspired approaches perceive the environment based on vision sensors.
However, vision sensors are usually very sensitive to the changes in lighting conditions and atmospheric effects of the environment, making them an unreliable solution to more advanced navigation tasks.
Light detection and ranging (\LiDAR) technology, on the other hand, can be more reliable as it is independent of ambient lighting conditions, and has higher accuracy and robustness than cameras~\cite{hiremath2014laser}.
A similar sensory principle with \LiDAR in nature can be found in bats, who emit brief ultrasound pulses and capture the echoes reflected from surrounding objects in a cave to perform localization.
Hence, it has significant potentials to investigate the biologically inspired localization using \LiDAR sensors.

In this work, we propose a biologically-inspired global localization system using a \LiDAR sensor that utilizes the hippocampal model and employs a landmark-based re-localization mechanism.
The proposed global localization system adapts and implements the pose cell model and the local view cell model,
which are originally inspired by the place cells and head direction cells in the hippocampus~\cite{milford2004ratslam}.
The models build a location representation in the space, and perform the localization iterations with the excitatory and inhibitory activity dynamics.
The landmark-based re-localization is leveraged to recovery from localization failures with the cues of familiar scenes.

Compared with conventional probabilistic localization methods, we propose the biologically inspired method by closely mimicking the navigation system of mammals and their re-localization strategy based on the sparse landmarks in the environment.
Therefore, the proposed method is free of probabilistic models that are used to model the uncertainties, such as the sensory noise.
The landmark-based re-localization enables the localization system to recovery from localization failures and solve the kidnapped robot problem faster and more efficiently compared to Monte Carlo localization.
Simulation and real-world experiments show that the proposed method can greatly reduce the localization errors by $40\%-50\%$ compared with MCL.
The experimental results also demonstrate the high applicability, and reliability of the proposed biologically-inspired localization system in different localization scenarios.
In addition, with the development of neuromorphic processors and hardware platforms~\cite{mead1990neuromorphic,indiveri2011neuromorphic}, the biologically-inspired navigation algorithms have the potential to implement and deploy on neuromorphic hardware to lower the computational costs~\cite{kreiser2018pose,davies2018loihi}.



\section{Related Work}

As Global Navigation Satellite System is a reliable source for global localization for the outdoor environment, there is a substantial body of research on multi-sensor fusion to improve the localization robustness based on GNNS~\cite{wan2018robust}.
Similarly, many studies on global localization for the indoor environment aim to provide an external reference system for global robot positioning, such as WLAN-based methods~\cite{zafari2019survey,torres2014ujiindoorloc,fang2011principal}.
These systems, however, usually require pre-deployment in the environment, and the runtime performance can be subject to signal interference.

A large amount of literature has investigated the global localization problem based on probabilistic methods, many of which have been successfully deployed on robots for outdoor and indoor scenarios~\cite{thrun2000probabilistic,dellaert1999monte,fox1999markov}.
Monte Carlo Localization (MCL)~\cite{dellaert1999monte} is one of the most commonly used global localization solutions for the indoor environment~\cite{thrun2002probabilistic}.
Proposed by Dellaert, Fox, et al. Monte Carlo Localization utilizes the particle filter to sample the posterior belief of the robot's position in the grid-based representation for the environment.
MCL is more efficient compared to grid-based localization~\cite{dellaert1999monte}, and there are existing studies focusing on improving the efficiency by enhancing the flexibility of particle sampling~\cite{fox2001kld}.
However, the performance of MCL is still subject to the number of sampling particles in the iterations, and one of the main disadvantages of MCL is that it can be hard for MCL to solve the kidnapped robot problem, recovering from the global localization failures~\cite{thrun2000probabilistic}.

There is a small body of literature with regard to global localization algorithms based on biological inspiration.
Siagian et al.~\cite{siagian2007biologically} proposed a visual localization system for outdoor environments based on Monte Carlo localization and a camera sensor, which focuses on the pre-processing of visual landmarks by mimicking the human visual capabilities of visual feature extraction.
The algorithm is still MCL-based, and the accuracy of the visual localization system can be diminished in indoor applications without abundant visual features.
There are rare studies that are concerned with biologically inspired localization in indoor environments based on \LiDAR sensor.


\section{Methodology}

\newcommand{\pose}{\mathbf{p}}

In this section, we first describe the architecture of the proposed method and present the model of each component, namely, the pose cell network and the local view cell module.

\subsection{Overview}

The proposed localization system consists of two main components: the pose cell network and the local view cell module.
The pose cell network mimics and implements the place cells and head direction cells in the hippocampus to build the location representation in the space, and also to perform the localization iterations with the excitatory and inhibitory activity dynamics.
In the localization iterations, the location hypotheses in the pose cells are being re-distributed based on the robot motions and sensor observations, and further converge to the best estimate.
The local view cell module is used to memorize and store a sparse set of landmarks during the process of mapping and real-time localization.
The local view cell module aims to activate the associated cell in the pose cell network when the robot revisits a recorded place, by which we can boost the localization progress and help recovery from localization failures.

In the framework of mobile robot localization, the global localization system aims to solve the problem of estimating the real-time pose of a robot on a given map of the environment based on the motion data and the sensory measurements from the robot, without or with rough knowledge of its initial position.
For the 2D localization problem for indoor environments,
the pose state of the robot is represented by a three-dimensional vector that describes the position and the heading direction: $ \pose = ({x}, {y}, \theta)^T$.

\subsection{Pose Cell Network}

\newcommand{\pc}{{\mathbf{PC}}}
\newcommand{\expNode}{{e}}
\newcommand{\link}{{l}}

\newcommand{\pcells}{{\pc_{x^\prime ,y^\prime, \theta^\prime}}}

\newcommand{\cellSizeXY}{{k_{xy}}}
\newcommand{\cellSizeTheta}{{k_{\theta}}}
\newcommand{\cellNumXY}{{N_{xy}}}
\newcommand{\cellNumTheta}{{N_{\theta}}}

\makeatletter
\renewcommand*\env@matrix[1][\arraystretch]{%
  \edef\arraystretch{#1}%
  \hskip -\arraycolsep
  \let\@ifnextchar\new@ifnextchar
  \array{*\c@MaxMatrixCols c}}
\makeatother

The pose cell network is a 3D continuous attractor network model, which was first presented by Milford et al. to solve the mapping problem~\cite{samsonovich1997path,milford2004ratslam}.
In this work, the pose cell network is adapted to maintain multiple location hypotheses in the space and perform real-time localization based on the sensor measurement input and the re-localization activities from the local view cells.
The confidences of the maintained location hypotheses are described as the activity levels of the pose cells.
The cells in the pose cell network are connected with their nearby cells with excitatory and inhibitory connections, which enable the location hypotheses to distribute and be filtered in the form of pose cell activities.

The pose cell network can be denoted as a 3D matrix of the cell activity levels: $\pc_{x^\prime ,y^\prime, \theta^\prime}, ~x^\prime ,y^\prime, \theta^\prime \in \mathbb{N}$.
Each of the three dimensions in the pose cell network
is a discrete representation for the three respective dimensions of the robot pose $\pose = (x, y, \theta)^T$ in $\mathbb{R}^3$. The mapping from the $\mathbb{R}^3$ space to the pose cell coordinates is done by the transformation described as follows:
\begin{equation}
	\begin{pmatrix}[1.6]
		x^\prime\\
		y^\prime\\
		\theta^\prime
	\end{pmatrix}
	= 
	\begin{pmatrix}[1.6]
		\floor*{x/\cellSizeXY+\cellNumXY/2}\\
		\floor*{y/\cellSizeXY+\cellNumXY/2}\\
		\floor*{\theta/\cellSizeTheta+\cellNumTheta/2}
	\end{pmatrix},
\end{equation}
where $\cellSizeXY$ and $\cellSizeTheta$ are the constant metric sizes of a single cell in the $x,y$ and $\theta$ dimensions, which can be also referred to as the resolution of the pose cells.
$\cellNumXY$ is the size in cells of the $ x^\prime $ and $ y^\prime $ dimensions of the pose cell network, and $ \cellNumTheta $ is the size in cells of the $ \theta^\prime $ dimension.
Therefore, a coordinate $(x, y, \theta)^T$ in $\mathbb{R}^3$ is scaled and shifted based on the resolution and the network size, to be represented by the pose cell network.
To cope with the loss of accuracy by the quantification, methods are proposed to alleviate the quantification effect, which will be detailed in the following subsections.

\subsubsection{Initial Estimate for Localization}

In the initial stage of the localization process, to start with, the pose cell network should to be initialized with an initial distribution of activity.
At the beginning of the global localization, without any knowledge of the robot's location, the robot can be placed in any position of the free space.
In the case with a given initial estimate with a pose $ \pose_{0} $ and a variance $ \sigma^{2} $, an initial activity distribution can be created by sampling $ n $ number of random cells $ \pcells $ from the Gaussian distribution $ \mathcal{N}(\pose_{0}, \, \sigma^2) $ and inject the selected cells with an initial amount of energy $ 1/n $.

Without a given initial estimate, the initialization of the pose cells can also be done with a higher variance $ \sigma^{2} $ and a higher sample number $ n $ to cover a larger range of space from the center of the pose cells, which can possibly cause a high resource overhead and performance issue.
However, with the help of the local view cells to directly activate the associated pose cells when a landmark is detected, the proposed localization system relies much less on initial estimates to fast start localization and recovery from localization failures than the conventional probabilistic localization algorithms.

\subsubsection{Path Integration}

When the robot moves, the pose cell network performs path integration~\cite{milford2004ratslam} to keep the track of the pose changes of the robot, which also enables the pose cells to update the localization estimates and subsequently re-distribute the activity based on the \LiDAR sensor observation.
Given a motion input from the odometry sensor on the robot, which can be denoted as a pose change $\Delta \pose=(\Delta x,\Delta y,\Delta \theta)^T$,
the activity update in the pose cells is defined as
\begin{equation}
\small
\begin{gathered}
	\Delta \pc_{x^\prime ,y^\prime, \theta^\prime} = 
	\sum^{\delta_{x}+1}_{i=\delta_{x}}
	\sum^{\delta_{y}+1}_{j=\delta_{y}}
	\sum^{\delta_{\theta}+1}_{k=\delta_{\theta}} {
		r_{ijk} \cdotp \pc_{x^\prime+i,\,y^\prime+j,\,\theta^\prime+k}
	}~, \\
	\delta_{x} = \lfloor \Delta x/\cellSizeXY \rfloor,
	\delta_{y} = \lfloor \Delta y/\cellSizeXY \rfloor,
	\delta_{\theta} = \lfloor \Delta \theta/\cellSizeTheta \rfloor,\\
	r_{ijk} = 
	(1 - |i - \Delta x/\cellSizeXY|)
	(1 - |j - \Delta y/\cellSizeXY|)
	(1 - |k - \Delta \theta/\cellSizeTheta|).
\end{gathered}
\end{equation}
For each pose cell $\pcells$, the activity is shifted based on the scaled movements $\delta_{x}$, $\delta_{y}$, and $\delta_{\theta}$ along three dimensions. In addition, to reduce the accuracy loss by quantification, the pose change is not quantified in the process of path integration. Instead, by taking the decimal parts of the scaled pose changes into account, for each dimension, the shifted activity will be spread in a unit of two adjacent cells proportionally based on the residual. With path integration, the pose cell network is able to track the movements of the robot. However, due to the noise in the odometry sensor, position errors can accumulate in this process. To cope with the problem, local excitation is introduced in the next subsection to help improve the accuracy and stability of the pose cell network.

\subsubsection{Attractor Dynamics}\label{sec:attractor-dynamics}

\newcommand{\globalInhibitionCoef}{{s_g}}
\newcommand{\localExcitationCoef}{{s_e}}

\newcommand{\gridMap}{{\mathbf{M}}}
\newcommand{\lidarMeasurement}{{z}}

\newcommand{\matchBoundary}{{\epsilon}}
\newcommand{\scanSize}{{N_\lidarMeasurement}}

In the proposed work, the 3D pose cell attractor network employs observation updating, excitatory activity, and inhibitory activity
to self-update the activity levels in the pose cells over time to perform the global localization.
The stable state of the pose cell network, in which the active cells are clustered as an activity packet, yields the localization estimate $\pose$ as the centroid of the activity packet.

On receiving the real-time observation from the \LiDAR sensor, the activity in the pose cells is updated 
to evaluate the confidence changes of the encoded location hypotheses based on the sensor measurement.
Given a \LiDAR scan measurement $\lidarMeasurement$ for time $t$, the new activity level for each active pose cell is computed as follows:
\begin{equation}
\begin{gathered}
	\pc_{x^\prime ,y^\prime, \theta^\prime}^{\,t+1} =
	p(\pose^\prime|\lidarMeasurement^t)
	\cdotp \pc_{x^\prime ,y^\prime, \theta^\prime}^{\,t},\\
	\pose^\prime=(x^\prime ,y^\prime, \theta^\prime)^T,
\end{gathered}
\end{equation}
where 
$p(\pose^\prime|\lidarMeasurement^t)$ is a conditional probability describing the likelihood of the location hypothesis $\pose^\prime$ for a pose cell matching the current \LiDAR observation $\lidarMeasurement^t$.
For a 2D \LiDAR, a complete scan measurement is a point cloud consisting of a series of endpoint distances with a static angle increment, and can be defined as a sequence $\lidarMeasurement =\{\langle d_i, \theta_i \rangle,~ i=1,...,\scanSize\}$,
in which $d_i$ is the distance for the $i$-th endpoint and $\theta_i$ is the corresponding scanning angle, while $\scanSize$ denotes the size of the \LiDAR measurement.
To evaluate the likelihood of a location hypothesis,
in this work, a given pre-built map for the environment is used to match with the \LiDAR observation.
The map $\gridMap$ is defined as a 2D occupancy grid map $\gridMap(x^\prime,y^\prime)$, which is the most common form of 2D maps used in indoor environments~\cite{thrun2002probabilistic,elfes1989using}.
Thus, given the map $\gridMap$ and the \LiDAR point cloud $\lidarMeasurement^t$, the likelihood for each active pose cell is computed as follows:
\begin{equation}
\small
\begin{gathered}
	p(\pose^\prime|\lidarMeasurement^t) = \frac{1}{\scanSize} \cdotp \sum^{\scanSize}_{i}
		\max_{\matchBoundary_1, \matchBoundary_2 \in \{-\frac{\cellSizeXY}{2},\frac{\cellSizeXY}{2}\}}{
			\gridMap\big(
				\lfloor x_i^\prime + \matchBoundary_1 \rfloor,
				\lfloor y_i^\prime + \matchBoundary_2 \rfloor
			\big)
		},\\
	x_i^\prime = x^\prime + d_i / \cellSizeXY \cdotp \cos \big(\theta_i+\cellSizeTheta(\theta^\prime-\cellNumTheta/2)\big),\\
	y_i^\prime = y^\prime + d_i / \cellSizeXY \cdotp \sin \big(\theta_i+\cellSizeTheta(\theta^\prime-\cellNumTheta/2)\big).
\end{gathered}
\end{equation}
In the process, the \LiDAR point cloud is first scaled and transformed into the pose cell space, in which the positions of the endpoints are shifted and rotated based on the position of the pose cell.
Note that to simplify the expression, we assume the map and the pose cell network share the same resolution $\cellSizeXY$.
With the given grid map $\gridMap$, the occupancies for the positions of the transformed endpoints are accumulated,
in which to alleviate the quantification effect for the \LiDAR point cloud and the pose cell network, the $2\times2$ adjacent grids in the map are taken into account in the matching process.
The aggregate matching occupancy then is normalized to evaluate how the current \LiDAR observation matches the map at the specific pose cell's position,
which can reflect the confidence of the location hypothesis,
by appling to the pose cell activity.

\newcommand{\gaussFunc}{{\mathcal{G}}}

After updating the activity in the pose cell based on the sensor observation,
the excitation and inhibition are introduced in the pose cell network to distribute and diminish the pose cell activity to complete the localization iteration.

To distribute and generate new activity around high confidence hypotheses,
the local excitation is performed to improve the localization accuracy.
The activity update in the process of local excitation is given as follows:
\begin{equation}
\small
\begin{split}
	\Delta \pc_{x^\prime ,y^\prime, \theta^\prime} = 
	\sum^{\cellNumXY}_{i}
	\sum^{\cellNumXY}_{j}
	\sum^{\cellNumTheta}_{k}
	{
		\gaussFunc_e(i-x^\prime, j-y^\prime, k-\theta^\prime) \cdotp \pc_{i ,j, k}
	}
\end{split}
\end{equation}
in which $\gaussFunc_e$ is a 3D Gaussian distribution to weigh the nearby activity levels based on the distances to them. Each active cell excites the nearby cells and distributes the pose cell activity. By exciting and reinforcing the cells around the active cells with higher localization confidences, local excitation re-distributes the location hypotheses which can reduce the accumulated errors introduced in path integration and the observation matching errors caused by the noise from the \LiDAR sensor, and further improves the stability and accuracy of the proposed localization system.

To stabilize the activity in the pose cell network over a long period of time, local inhibition is involved to curb the activity expansion entailed by local excitation.
Similarly, local inhibition is performed in a negative form of local excitation with a different Gaussian distribution variance and coefficient. Therefore, each cell inhibits near cells to stabilize the activity packet~\cite{samsonovich1997path,milford2004ratslam}, and make cells with a low activity level be less and less active over time, which leads to the location hypotheses with lower confidence being able to be diminished and filtered based on the \LiDAR observations.
Note that by setting the variances and coefficients, the local excitation is tuned to be slightly dominant in this process, in order to ensure the cell activity is able to distribute when the average location confidence is low, instead of falling into local optima.

Besides local inhibition, global inhibition is performed
to introduce a fix-rate slow decay for all the active cells in the pose cell network,
in which the activity update is defined as follows:
\begin{equation}
	\Delta \pc_{x^\prime ,y^\prime, \theta^\prime} = - \min(\pc_{x^\prime ,y^\prime, \theta^\prime}, \globalInhibitionCoef),
\end{equation}
where $\globalInhibitionCoef$ is a constant factor as the decay rate in each iteration.
Therefore, the activity of each active pose cell slightly decays in every localization iteration, which enables the pose cell network to diminish and clear insignificant but non-zero hypotheses, to improve the performance of the localization system over a long period of time.
After local and global inhibition, activity normalization among the whole pose cell network is performed to scale and normalize the activity of each active cell to $[0,1]$ based on its weight on all active pose cells. Normalization at the end of the localization iteration ensures pose cells with high activity levels will not be affected by the inhibition over a long period of time.

With path integration, observation updating, activity excitation, and inhibition, the proposed localization system is able to complete the localization iteration based on the motion data and \LiDAR observations from the robot.
When the activity in the pose cell network converges as a single activity packet,
the averaged centroid of the dominant activity packet is identified to be the best pose estimate for the robot.

\subsection{Local View Cells}

\newcommand{\conMatrix}{{\mathbf{A}}}
\newcommand{\viewCells}{{V}}
\newcommand{\localViewInjectionCoef}{{s_v}}

The local view cell module is adapted to extract and maintain landmarks from the \LiDAR observations in the proceeding mapping process and the global localization process,
and associate the landmarks with the pose cells to perform re-localization when the landmarks are detected in the process of global localization.

Given a \LiDAR point cloud $\lidarMeasurement =\{\langle d_i, \theta_i \rangle,~ i=1,...,\scanSize\}$ from the sensor measurement,
the local view cell module first downsamples and quantifies the point cloud to reduce the data scale and also diminish the noise effect from the \LiDAR sensor.
With the pre-processed \LiDAR point cloud, a landmark is generated by extracting the corner points in the point cloud as the geometry feature of the current \LiDAR observation, which reduces the complexity of the landmarks and enables the local view cell module to do the fast landmark detection.
The generated landmark, which consists of a sparse set of key points, is then registered as a new local view cell $\viewCells_i$ and
associated with the current state of the pose cell network,
in which the connectivity between the local view cells and the pose cells is defined as a sparse adjacency matrix $\conMatrix$.
In the process of global localization, when a landmark is detected, the corresponding local view cell is therefore activated,
injecting a re-localization activity into the pose cell network to provide a location hypothesis.
The activity of the associated pose cell is updated based on the activity of local view cells $\viewCells$, which can be described as follows:
\begin{equation}
	\Delta \pc_{x^\prime ,y^\prime, \theta^\prime} = 
	\localViewInjectionCoef \cdot \sum_{i}{\conMatrix_{i, x^\prime ,y^\prime, \theta^\prime} \cdot \viewCells_i},
\end{equation}
where $\localViewInjectionCoef$ is the constant coefficient for the activity injection.
In the process, for all the local view cells, the activity is injected only if the local view cell $\viewCells_i$ is active and there is a connection $\conMatrix_{i, x^\prime ,y^\prime, \theta^\prime}$ from $\viewCells_i$ to the pose cell $\pc_{x^\prime ,y^\prime, \theta^\prime}$.

Given the \LiDAR landmark cues from the local view cells, the pose cell network is able to generate and distribute more location hypotheses when the average localization confidence is low.
Due to the potential landmark ambiguity, the activity from the local view cells is not injected as a dominant activity packet in the pose cell network, but to provide a location hypothesis to compete with the others.
The landmark-based re-localization enables the proposed localization system to
boost the localization progress at the beginning without the help of initial estimates,
and fast recovery from localization failures, such as the \emph{kidnapped robot problem}, in which the localization fails to keep the track of the robot's location because the robot is arbitrarily moved to another location~\cite{engelson1992error}.


\section{Experiments}

To evaluate the performance of the proposed localization system, experiments in a simulator and a real-world indoor environment were conducted.
The localization has been tested in comparison with the Monte Carlo localization (MCL) algorithm~\cite{dellaert1999monte}, which is commonly used for indoor scenarios with \LiDAR sensors~\cite{fox2001kld}.
In the experiments, we evaluate the results of global localization with and without initial estimates, and the performance of recovery from artificial localization failure situations as known as the kidnapped robot problem, to show the effectiveness of the proposed method and its capability of generalizing to different localization scenarios.

\subsection{Experimental Setup}

\begin{figure}[t]
	\vspace{-7pt}
	\centering
		\subfloat[Simulation maze scheme]{
		\includegraphics[width=0.45\linewidth]{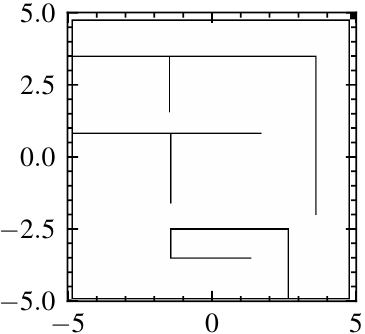}
		\label{fig:sim-maze}
		}\hfil
		\subfloat[Simulation experiment trajectory]{
		\includegraphics[width=0.45\linewidth]{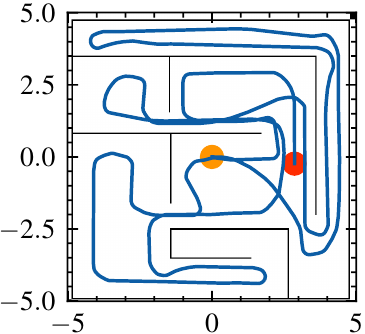}
		\label{fig:sim-trajectory}
		}\\
		\subfloat[Real-world maze scheme]{
		\includegraphics[width=0.45\linewidth]{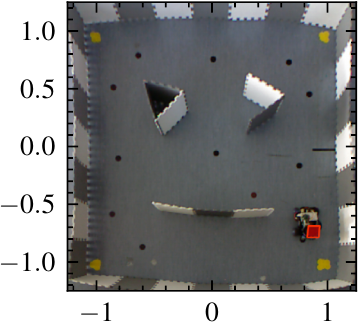}
		\label{fig:garching-maze}
		}\hfil
		\subfloat[Real-world experiment trajectory]{
		\includegraphics[width=0.45\linewidth]{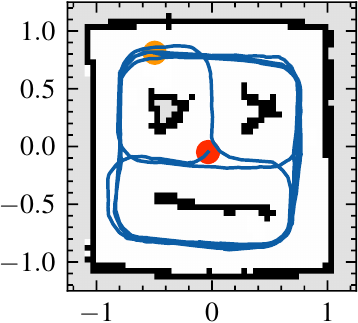}
		\label{fig:garching-trajectory}
		}
	\caption{Schemes and trajectories for simulation and indoor experiments}
	\label{fig:maze}
	\vspace{-1pt}
\end{figure}


To provide simple and configurable environments to test the proposed localization method, we set up a simulated scene based on the Robot Operating System framework~\cite{quigley2009ros}, as shown in \figurename{\,\ref{fig:sim-maze}}.
The maze-like environment consists of a collection of identifiable geometry features and symmetrical scenes, which can be challenging to the localization algorithms with the environment ambiguities.
The trajectory shown in \figurename{\,\ref{fig:sim-trajectory}} reflects the robot's movements during the process of the simulation experiments. The robot was manually operated to explore the environment, starting from the position of the yellow marker and ending at the red marker.

To examine the practical capability of the proposed method, we also construct a real-world maze environment.
\figurename{\,\ref{fig:garching-maze}} show the maze scheme and environment settings for the indoor scenario.
In the real-world indoor experiments, a mobile robot platform was remotely controlled to navigate the environment, which is equipped with an embedded IMU sensor and a 2D \LiDAR sensor.
To provide a localization ground truth, a bird view camera is used to acquire the position of the agent in real-time, which will be subsequently utilized to evaluate the localization accuracy.

\subsection{Localization Results}

\begin{figure}[t]
	\centering
	\smallskip
		\includegraphics[width=0.95\linewidth]{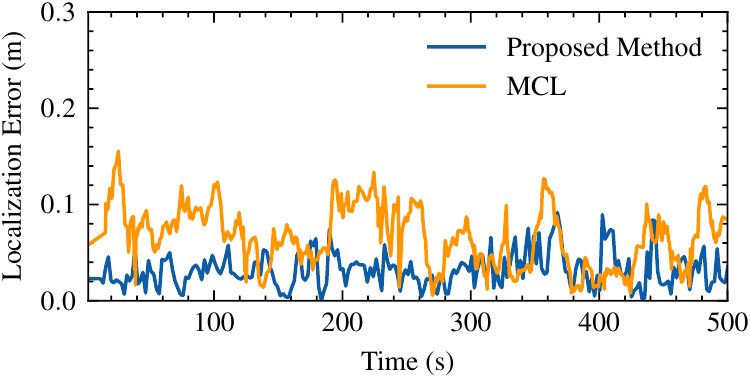}
	\caption{Localization errors in the simulation experiments}
	\label{fig:error_sim}
\end{figure}
\begin{figure}[t]
	\centering
		\includegraphics[width=0.95\linewidth]{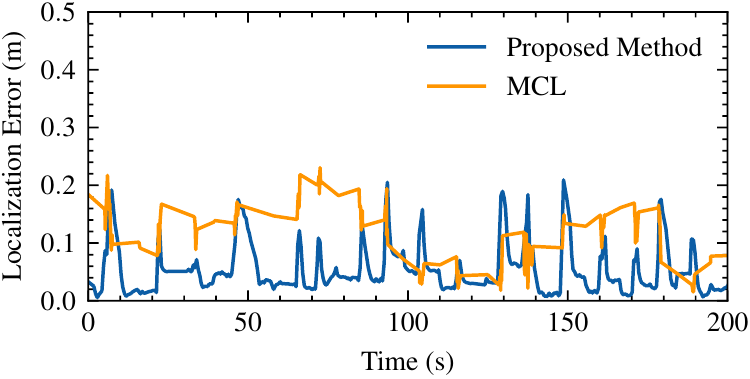}
	\caption{Localization errors in the indoor experiments}
	\label{fig:error_garching}
	\vspace{-12pt}
\end{figure}

\begin{table}[b]
\vspace{-12pt}
\renewcommand{\arraystretch}{1.3}
\setlength\tabcolsep{3pt}
\caption{Average Localization Errors}
\label{table:errors}
\centering
\begin{tabular}{c||c|c|c|c}
	\hline
	Simulation & X & Y & Distance & RMSE \\
	\hline
	\bfseries \makecell{Proposed Method} & \bfseries 0.0192\,m & \bfseries 0.0215\,m & \bfseries 0.0319\,m & \bfseries 0.0366\,m\\
	\hline
	MCL & \SI{0.0471}{m} & \SI{0.0364}{m} & \SI{0.065}{m} & \SI{0.0720}{m} \\
	\hline\hline
	
	\makecell{Indoor Scenario} & X & Y & Distance & RMSE \\
	\hline
	\bfseries \makecell{Proposed Method} & \bfseries 0.0319\,m & \bfseries 0.0337\,m & \bfseries 0.0532\,m & \bfseries 0.0681\,m \\
	\hline
	MCL & \SI{0.0672}{m} & \SI{0.0612}{m} & \SI{0.1033}{m} & \SI{0.1144}{m} \\
	\hline
\end{tabular}
\end{table}

To evaluate the effectiveness of the proposed localization system, we first report the accuracy of the localization results.
\figurename{\,\ref{fig:error_sim} reveals the absolute translation errors over time in the localization process of the simulation experiments. As shown in the figure, the proposed biologically-inspired method has a lower level of localization errors during the experiment process in comparison with the conventional MCL algorithm.
The average distance error and the root-mean-square error (RMSE) of the proposed method are
\SI{0.0319}{m} and \SI{0.0366}{m}.
The errors are reduced by $50.9\%$ and $49.2\%$ respectively
from the average errors of MCL
which are \SI{0.065}{m} and \SI{0.0720}{m}.
The results show that the proposed method is more accurate and stabler compared to MCL in the simulation.

\figurename{\,\ref{fig:error_garching}} plots the localization errors of the experiments in the indoor maze scenario.
Due to the limitation of the field of view and the installation height of the bird view camera which provides localization ground truth in the experiments, the maze setting in the indoor experiments has a more constrained size in comparison with the simulation experiments.
However, the scenario can be more challenging, in which the localization algorithms are more sensitive to the measurement noise and latency from sensors.
The localization errors in \figurename{\,\ref{fig:error_garching}} for both the proposed localization system and MCL are slightly higher than that in the simulation experiments due to the complexity of the realistic environment, partial ambiguities in the maze setting, and the indeterministic noise and latency of the imperfect IMU and \LiDAR sensor.
However, as shown in the graph, the proposed biologically-inspired method still outperformed MCL with lower average errors in the localization process.
Compared with the result of MCL, the average distance error of the proposed method is reduced by $48.5\%$ from \SI{0.1033}{m} to \SI{0.0532}{m}, and the RMSE is reduced by $40.5\%$ from \SI{0.1144}{m} to \SI{0.0681}{m}.
In the curve,
the error surged many times due to the repetitive stalled status and fast starts of the robot.
However, the spikes in the error curve reveal that
the proposed method has the capability of fast correcting and recovering the cumulative localization errors introduced by the sensor noise and latency.

In \tablename{\,\ref{table:errors}} we list the average errors in the simulation and indoor experiments, including the translational errors in the \emph{x} and \emph{y} axes respectively, absolute distance errors, and the root-mean-square errors (RMSE). The proposed method has outperformed the conventional MCL algorithm with lower errors in all the evaluated terms, which means it has not only higher accuracy but also higher stability.

\begin{figure}[t]
	\centering
	\smallskip
		\includegraphics[width=0.95\linewidth]{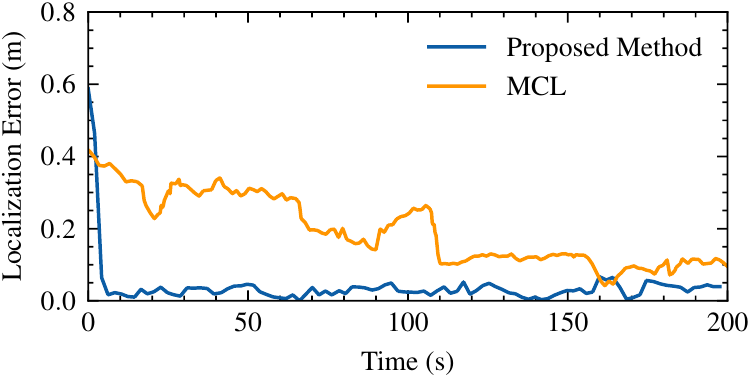}
	\caption{Localization errors without initial estimates}
	\label{fig:error_no_init_sim}
	\vspace{-12pt}
\end{figure}

In the experiments, we also evaluate the global localization algorithms without the initial estimates.
Global localization without initial estimates can entail the localization system a higher workload with a larger initial distribution of location hypotheses. \figurename{\,\ref{fig:error_no_init_sim}} illustrates the localization process for an experiment without initial estimates and the proposed re-localization, which will be assessed in the following subsection.
The result reveals that in the localization process the proposed method fast produced an accurate pose estimate as the error dropped dramatically to a low level in \SI{7}{s} at the beginning. In comparison, the re-distribution of the particles in MCL is driven by the robot motions, resulting in the slow decline of the estimation error in the localization process.

The localization results show that the proposed biologically inspired method with the pose cell network is capable of tracking the robot movements, fast distributing and updating location hypotheses, and yielding accurate pose estimates for the global localization process.
In addition, the results also reflect that the proposed method can well generalize to real-world indoor environments.


\subsection{Recovery from Artificial Localization Failures}

\begin{figure}[t]
	\centering
	\smallskip
		\includegraphics[width=0.95\linewidth]{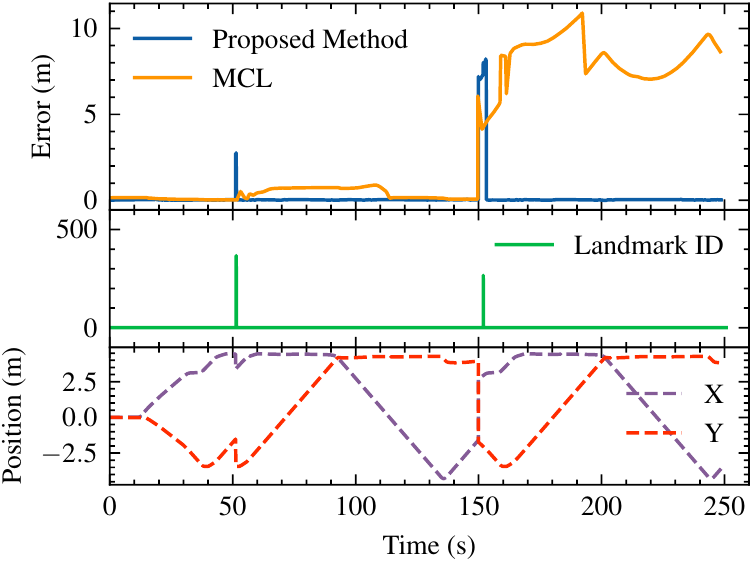}
	\caption{Localization recovery from artificial failures}
	\label{fig:kidnapping}
	\vspace{-12pt}
\end{figure}


In this subsection, we assess the landmark-based re-localization approach of the proposed method by conducting localization recovery experiments in the situation of artificial failures.
\figurename{\,\ref{fig:kidnapping}} portrays the results of two complete test cases of localization failure recovery.
As shown in the graph, around \SI{50}{s}, a short-distance robot kidnapping was enforced by instantly moving the robot to a near place, which can be identified in the figure with the instant position change in the \emph{x}, \emph{y} curves.
The second artificial failure was introduced around \SI{150}{s} by re-positioning the robot at a place far away from its prior position.
The localization errors in the upper part of the graph show that the MCL algorithm recovered from the first short-distance failure with a long re-localization process which lasted for \SI{60}{s}, and it was not able to re-localize the robot in the second long-distance artificial failure.
The landmark curve reflects the activity injections from the local view cell module to the pose cell network when localization failures happened, in which two landmarks were detected and the activity was injected around \SI{50}{s} and \SI{150}{s}.
The fast dropping spikes in the error curve of the proposed method show that the robot was successfully re-localized from both the failures of short and long-distance, with much less time on the re-localization process.
The result demonstrates that
with the proposed landmark-based re-localization the system is more reliable and capable of swiftly recovering from localization failures and solve the kidnapped robot problem.


\section{Conclusion}

In this paper, we present a biologically-inspired global localization system using a \LiDAR sensor that consists of the pose cell network and the local view cell module.
The pose cell network mimics the place cells and head direction cells in the hippocampus to build a spatial representation and maintain, distribute, and filter the location hypotheses with the excitatory and inhibitory activity dynamics. 
The local view cell module memorizes and store landmarks, and activate the associated pose cell when the identical landmark is detected, which can help recovery from localization failures, and solve the kidnapped robot problem.
The experimental results demonstrate the high accuracy, applicability, and reliability of the proposed biologically-inspired localization system in different localization scenarios.

\newpage

\IEEEtriggeratref{14}
\bibliographystyle{IEEEtran}
\bibliography{ref}

\end{document}